\documentclass[11pt]{article}
\usepackage{coling2018}
\usepackage{times}
\usepackage{url}
\usepackage{latexsym}
\usepackage{tabularx}
\usepackage{multirow}
\usepackage{graphicx}
\usepackage{xcolor}
\usepackage{textcomp}
\usepackage[T4,T1]{fontenc}
\newenvironment{tfour}{\fontencoding{T4}\selectfont}{}
\usepackage{newunicodechar}
\usepackage{enumitem}

\newunicodechar{ɔ}{\fon}

\title{FFR v1.1: Fon-French Neural Machine Translation}
\author{Bonaventure F. P. Dossou \\
 Kazan Federal University\\
\texttt{femipancrace.dossou@gmail.com} \\\And
Chris C. Emezue \\
Kazan Federal University\\
\texttt{chris.emezue@gmail.com} \\}

\begin{document}
\maketitle
\begin{abstract}
All over the world and especially in Africa, researchers are putting efforts into building Neural Machine Translation (NMT) systems to help tackle the language barriers in Africa, a continent of over 2000 different languages. However, the low-resourceness, diacritical, and tonal complexities of African languages are major issues being faced. The FFR project is a major step towards creating a robust translation model from Fon, a very low-resource and tonal language, to French, for research and public use. In this paper, we introduce FFR Dataset, a corpus of Fon-to-French translations, describe the diacritical encoding process, and introduce our FFR v1.1 model, trained on the dataset. The dataset and model are made publicly available at \url{https://github.com/bonaventuredossou/ffr-v1}, to promote collaboration and reproducibility.%and are still being improved daily.
\end{abstract}
\section{The FFR Dataset:  FFR1 and FFR2}
\label{ffrdatasets}
The FFR Dataset is a project to compile a large, growing corpus of carefully cleaned of Fon - French (FFR) parallel sentences for machine translation, and other NLP research-related, projects \cite{ffr}. There are currently two versions of the FFR dataset: the initial FFR dataset (FFR1) and the latest version (FFR2).

The major sources for the creation of FFR1 were JW300 \cite{agic-vulic-2019-jw300} and BeninLangues\footnote{https://beninlangues.com/} with 27980 and 89049 aligned sentences respectively, giving a total of 117,029 parallel sentences. JW300 (JW) contains translations of Jehovah Witness sermons in over 100 languages, while BeninLangues (BL) contains vocabulary words, short expressions, small sentences, complex sentences, proverbs, as well as books of the Bible (Genesis 1 - Psalm 79). 

The initial samples contained various grammatical errors, incorrect and incomplete translations, which were disregarded by standard, rule-based cleaning techniques\footnote{Using Python Regex and String packages (https://docs.python.org/3/library/re.html) and NLTK preprocessing library (https://www.nltk.org/)}. FFR2, obtained after re-evaluation of translations in FFR1 by FFR natives, reduces JW and BL original samples respectively to 26510 and 27465 Fon-French parallel sentences. We also created a data statement \cite{bender-friedman-2018-data} for FFR2, which serves to help give a thorough overview of the dataset. Our data statement can be accessed at \url{https://github.com/bonaventuredossou/ffr-v1/blob/master/FFR-Dataset/Data_Statement_FFR_Dataset.pdf}. The tabular analyses shown in Table \ref{ffrdataset} below serve to give an idea of the range of word lengths for the sentences in FFR1 and FFR2. The maximum number of words-per-sentence for the Fon sentences, $max-fon$, is 109, for FFR1, and 88, for FFR2. That of the French sentences, $max-fr$, is 111 for FFR1 and 76 for FFR2. Therefore, the dataset (both FFR1 and FFR2) has a good range of short, medium and long sentences.
\begin{table}[h]
\caption{ \bf Analysis of Sentences in FFR1}
\label{ffrdataset}
\begin{center}

\begin{tabular}{lllll}
\multicolumn{1}{l}{} &\multicolumn{2}{l}{\bf FFR1} & \multicolumn{2}{l}{\bf FFR2} \\
\hline
\multicolumn{1}{l}{} &\multicolumn{1}{l}{\bf FON} & \multicolumn{1}{l}{\bf FRENCH}&\multicolumn{1}{l}{\bf FON} & \multicolumn{1}{l}{\bf FRENCH} \\
\# Very Short sentences [1-5 words]      &64301          &64255&27470          &30817        \\
\# Short sentences [6-10 words]            &13848                    &17183  &6898                    &12500  \\
\# Medium sentences [11-30 words]             &29113    &29857 &17529    &10582 \\
\# Long sentences [31-($max-fon$ or $max-fr$ )]          &9767           &5734 &2078           &76 \\
Total & \bf 117029 && \bf 53975&

\end{tabular}
\end{center}
\end{table}

%\begin{table}[h]
%\caption{ \bf Analysis of Sentences in FFR2}
%\label{cffrdataset}
%\begin{center}
%
%\begin{tabular}{lll}
%\multicolumn{1}{l}{} &\multicolumn{1}{l}{\bf FON} & %\multicolumn{1}{l}{\bf FRENCH} \\
%\# Very Short sentences [1-5 words]      &27470          %&30817        \\
% \# Short sentences [6-10 words]            &6898        %            &12500   \\
%\# Medium sentences [11-30 words]             &17529    %&10582 \\
%\# Long sentences [31-($max-fon$ or $max-fr$ )]          %&2078           &76 \\
%\end{tabular}
%\end{center}
%\end{table}

%\begin{table}[h]
%\caption{\bf Table showing how the different diacritics affect %the meaning of the Fon word \textit{to}.}
%\label{fonto}
%\begin{center}
%\begin{tabular}{ll}
%\multicolumn{1}{l}{\bf Words with accents} & %\multicolumn{1}{l}{\bf Meaning} \\
%tó & ears \\
%tò & sea  \\
%tô & country \\
%{t\begin{tfour}\m{o}\end{tfour}}&father \\
%\end{tabular}
%\end{center}
%\end{table}
\section{Data Preprocessing}
\label{dateprocess}
Initial analysis of Fon sentences revealed that different accents (or diacritics or tone marking)\footnote{\url{https://en.wikipedia.org/wiki/Fon_language#Tone_marking}} on same words affected their meanings, making it necessary to keep the accents (diacritics) of Fon tokens (words, characters). The importance of encoding diacritics (Diacritical Encoding (DE)) of African languages to NMT has been highlighted by researchers \cite{Orife2020ImprovingYD}, who in their experiments affirmed that DE reduces lexical disambiguation, and helps provide more morphological information to the model. DE was performed using the Normalization Form Canonical Composition (NFC) instead of the Normalization Form Canonical Decomposition (NFD) \footnote{\url{https://unicode.org/reports/tr15/#Norm_Forms}}. With NFC, characters are decomposed and then recomposed by canonical equivalence, while with NFD, they are simply decomposed by canonical equivalence, which removes all accents of Fon tokens. For example, considering the Fon word, \textbf{to}, with its different diacritical meanings, [(t$\acute{o}$,ears),(t$\grave{o}$,sea), (t$\hat{o}$, country), ({t\begin{tfour}\m{o}\end{tfour}},father)], we see that using NFC keeps the diacritics and consequently the meaning of the words, while using NFD, simply gives the word \textit{to} leading to ambiguities in the translation. 

%Considering the sentence \colorbox{pink}{xota ðò akpà {\begin{tfour}\m{o}\end{tfour}} jí é etɛwu mǐ ðó n{\begin{tfour}\m{o}\end{tfour}} ð{\begin{tfour}\m{o}\end{tfour}} nugbǒ}, the NFC result is \colorbox{pink}{xota ðò akpà {\begin{tfour}\m{o}\end{tfour}} jí é etɛwu mǐ ðó n{\begin{tfour}\m{o}\end{tfour}} ð{\begin{tfour}\m{o}\end{tfour}} nugbǒ} i.e. the exact sentence while NFD will produce \colorbox{pink}{xota ðo akpa {\begin{tfour}\m{o}\end{tfour}} ji e etɛwu mi ðo n{\begin{tfour}\m{o}\end{tfour}} ð{\begin{tfour}\m{o}\end{tfour}} nugbo}.%

\section{FFR v1.1 Model Structure and Training}
For our experiments, we used FFR2, described in section \ref{ffrdatasets}, which is an improvement of FFR1. We derived 43719, 4858 and 5398 training, validation and testing samples accordingly. We used the Tensorflow TextTokenizer\footnote{\url{https://www.tensorflow.org/api_docs/python/tf/keras/preprocessing/text}} with \textit{none} filter to tokenize FFR sentences and build the vocabularies (for Fon and French), from which numerical sequences or representations of each FFR sentence pair are built with the Tensorflow Preprocessing package\footnote{\url{https://www.tensorflow.org/api_docs/python/tf/keras/preprocessing/sequence}}, and used to train the model.

The FFR v1.1  model, like the FFR v1.0 \cite{ffr}, is based on the encoder-decoder configuration \cite{NIPS20145346,brownlee,tensorflow}. The encoders and decoders are made up of 128-dimensional gated rectified units (GRUs) recurrent layers \cite{lstm}, with a word embedding layer of dimension 512. A 30-dimensional attention model \cite{NIPS20145346,attention,lamba} was also applied in order to help the model make contextual and correct translations. The code for the model has been open-sourced at \url{https://github.com/bonaventuredossou/ffr-v1/blob/master/model_train_test/fon_fr.py}, to promote reproducibility and similar recent initiaves on machine translation of African languages like \cite{jade,masakhane}. FFR v1.1 model was trained using the Tensorflow v1.14 package \cite{tensorflow}.

\section{Initial Results and Findings}
We evaluated the FFR v1.1 model performance using BLEU \cite{papineni-etal-2002-bleu}, and GLEU \cite{gleu} metrics. GLEU, is a sentence-level evaluation metric similar to BLEU. 
\begin{table}[h]
\caption{Evlauation scores on test data}
\label{testscore}
    \centering
\begin{tabular}{lllll}

%\caption{BLEU and GLEU Scores on Test Data}
\multicolumn{1}{l}{} &\multicolumn{2}{l}{\bf FFR1 } & \multicolumn{2}{l}{\bf FFR2} \\
\hline

\multicolumn{1}{l}{} &\multicolumn{1}{l}{\bf BLEU} & \multicolumn{1}{l}{\bf GLEU} &\multicolumn{1}{l}{\bf BLEU} & \multicolumn{1}{l}{\bf GLEU}\\
Without DE&24.53&13.0&27.80&17.05  \\
With DE&\bf{30.55}&\bf{18.18} &\bf{ 37.15}&\bf{20.85} \\

\end{tabular}
\end{table}
As shown on Table \ref{testscore}, the FFR model, trained on both FFR1 and FFR2 showed an improvement when trained with DE. %The FFR v1.1 model performed better on FFR2 than on FFR1, and this testifies of the effectiveness of the deeper cleaning that has been made.

Table~\ref{sample-table} shows translations of interest from the FFR model sources from FFR2, illustrating the difficulty of predicting Fon words which bear different meanings with different accents. While the model predicted well for \#0 and \#1, it misplaced the meanings for \#2 and \#3.  
\begin{table}[h!]
\caption{Sample predictions and scores}
\label{sample-table}
\begin{center}
 \resizebox{\textwidth}{!}{
\begin{tabular}{ |c|c|c|c|c|c|c| } 
 \hline
\bf ID & \bf 0 &\bf 1 &\bf2&\bf3&\bf4&\bf5\\
 \textcolor{blue}{Source} & \textcolor{blue}{y$\acute{i}$ bo wa} & \textcolor{blue}{yi bo wa}  &\textcolor{blue}{h\begin{tfour}\m{o}\end{tfour}n} &\textcolor{blue}{h\begin{tfour}\m{o}\end{tfour}n} & \textcolor{blue}{s\'a amas\'in d\u o w\u u } &\textcolor{blue}{gb\begin{tfour}\m{e}\end{tfour}}
 \\ 
 \textcolor{brown}{Target} &  \textcolor{brown}{prends et viens} &  \textcolor{brown}{va et viens} & \textcolor{brown}{porte} & \textcolor{brown}{fuire} & \textcolor{brown}{oindre avec un m\'edicament} &\textcolor{brown}{pousser de nouvelles feuilles}  \\
 FFR v1.0 Model & prends et viens & va viens & scorpion &porte &se masser avec le remede &esprit de la vie\\
 BLEU/\bf CMS score & 1.0 & 1.0 & 0.0   &0.0 &0.0/\textbf{0.95} &0.25 /\textbf{ 0.9} \\ 
 \hline
\end{tabular} 
}
\end{center}
\end{table}
\subsection{The Context-Meaning-Similarity (CMS) metric}
Researchers have shown that automatic metrics are not necessarily a good substitute for human assessments of translation quality \cite{Turian03evaluationof,callison-burch-etal-2006-evaluating,graham}, due to issues like lexical-vs-semantic similarity and existence of many possible valid translations for each source sentence \cite{koehn,lo-etal-2013-improving,graham}.
During our experiments, we discovered that the FFR v1.1 model was able to provide predictions that were, although different from the target, similar in context to the target, as seen in sentence \#4. Both \textcolor{brown}{oindre avec un m\'edicament} and \textcolor{brown}{se masser avec le remede} convey the same idea in the context of the source sentence, \textcolor{blue}{s\'a amas\'in d\u o w\u u }.\newline
%Most African languages, due to the , have many dialects Also, there is the problem of bias in human assessment of translations in some African languages, due to the multi-ethnicity and varied dialects of the same language.

This led us to experiment a method we call CMS metric:
\begin{footnotesize}
\begin{enumerate}[noitemsep]
\item A subset of the testing data, consisting of 100 specially selected source, target and predicted sentences, was sent to five FFR natives.
\item \label{tscore} They were first given the source and prediction sentences and asked to give a score, $t \in [0,1]$, on how similar the source and prediction sentences were contextually. Note that this scoring was done with no knowledge of the reference, but through the innate experience of the native speakers.
\item  Then they were given the source and prediction along with the reference sentences and, simillar to step \ref{tscore} above, were instructed to give a score $t_{r}$.
\item Using a parameter, $\alpha$, we calculated the total score $t_{total} = \alpha*t + (1-\alpha)*t_{r}$. This parameter controls the tradeoff between the review of the prediction, when viewed on its own, and that of the prediction when viewed in contextual comaprison to the reference sentence. For our experiment, we set $\alpha = 0.7$, putting more weight on the prediction without the reference comparison.
\item The average of these scores was taken as the CMS score for each of the model\textquotesingle s predictions as given in sentence \#4 in Table~\ref{sample-table}. 
\end{enumerate}
\end{footnotesize}
An interesting feature of the CMS metric is the tradeoff, $\alpha$, which is especially useful for translation assessments in languages that have many dialects (like most African languages) and expressions with various possible contexts (like Fon). 
\section{Conclusion, Future Work and Acknowledgements}
In this paper, we introduced the creation of the FFR dataset: a corpus of Fon-French parallel sentences. We further trained an NMT system, and evaluated the translation quality using both the BLEU metric and our proposed CMS metric. Our project is at the pilot stage and therefore, there is headroom to be explored with the tuning of different architectures, learning schemes, transfer learning, tokenization methods for the FFR project (FFR Dataset, FFR model) improvement. Specifically, we are looking into leveraging monolingual data, encoding with subword units \cite{DBLP:journals/corr/SennrichHB15}, exploring data augmentation for low-resource NMT \cite{dataaugmentation,dataaugmentation1}, and training on a state-of-the-art Transformer model \cite{Vaswani2017AttentionIA}. We owe great thanks to Julia Kreutzer, Jade Abott and the Masakhane Community for their mentorship. We would also like to thank the FFR natives for the good translation services provided.

\blfootnote{
    %
    % for review submission
    %
    \hspace{-0.65cm}  % space normally used by the marker
    % Place licence statement here for the camera-ready version. See
    %Section~\ref{licence} of the instructions for preparing a
    %manuscript.
    %
    % % final paper: en-uk version 
    %
    This work is licensed under a Creative Commons 
    Attribution 4.0 International Licence.
    Licence details:
    \url{http://creativecommons.org/licenses/by/4.0/}.
    % 
    % % final paper: en-us version 
    %
    % \hspace{-0.65cm}  % space normally used by the marker
    % This work is licensed under a Creative Commons 
    % Attribution 4.0 International License.
    % License details:
    % \url{http://creativecommons.org/licenses/by/4.0/}.
}

%\bibliography{coling2018}
\bibliographystyle{acl}
\bibliography{winlp2020}

\end{document}